\documentclass[11pt]{article}

\usepackage[T1]{fontenc}
\usepackage{lmodern}
\usepackage[english]{babel}
\usepackage{microtype}
\usepackage[a4paper,margin=0.85in]{geometry}
\usepackage{graphicx}
\usepackage{booktabs}
\usepackage{tabularx}
\usepackage{multirow}
\usepackage{array}
\usepackage{siunitx}
\usepackage{amsmath}
\usepackage{enumitem}
\usepackage{caption}
\usepackage{subcaption}
\usepackage{float}
\usepackage{placeins}
\usepackage{xurl}
\usepackage[hidelinks]{hyperref}
\usepackage[nameinlink,noabbrev]{cleveref}

\hypersetup{
  pdftitle={Comparative Analysis of Lightweight CNNs for Resource-Constrained Devices: Predictive Performance, Efficiency Trade-offs, and Initialization Effects},
  pdfauthor={Tasnim Shahriar}
}

\graphicspath{{figures/}}
\setlist{nosep,leftmargin=*}
\newcolumntype{Y}{>{\centering\arraybackslash}X}

\title{Comparative Analysis of Lightweight CNNs for\\Resource-Constrained Devices:\\Predictive Performance, Efficiency Trade-offs,\\and Initialization Effects}
\author{Tasnim Shahriar\\{\itshape Independent Researcher}}
\date{}

\begin{document}
\maketitle

\begin{center}
\small
\textit{This manuscript is currently under consideration at a peer reviewed venue. The preprint may be updated following feedback received during the review process.}
\end{center}

\begin{abstract}
Lightweight convolutional neural networks are often compared using results obtained with different training recipes, input settings, and pretrained checkpoints. Such differences make architecture rankings difficult to interpret. This study presents a controlled benchmark of seven established CNNs across CIFAR-10, CIFAR-100, and Tiny ImageNet under a shared fine tuning protocol. The evaluation reports top-1 accuracy, macro F1, top-5 accuracy, parameter count, FP32 parameter storage, and multiply accumulate operations. EfficientNetV2-S records the highest observed top-1 accuracy on all three datasets, reaching 97.57\%, 86.98\%, and 78.73\%. EfficientNet-B0 remains within 0.85 percentage points of EfficientNetV2-S across the three datasets while requiring only about 21\% of its parameters and 14\% of its multiply accumulate operations on Tiny ImageNet. It therefore offers a favorable general balance between predictive performance and computational demand. MobileNetV3-Small is a strong candidate for ultra low resource settings. It uses about 40\% of the parameters and 15\% of the multiply accumulate operations of EfficientNet-B0 while retaining competitive accuracy. A matched comparison of ImageNet pretrained and randomly initialized EfficientNet-B0 and MobileNetV3-Small models shows that the pretrained advantage is substantially larger on CIFAR-100 and Tiny ImageNet than on CIFAR-10 under the fixed protocol. The results provide a focused reference for selecting established lightweight CNNs when predictive quality, parameter storage, and theoretical computation must be considered together.
\end{abstract}


\section{Introduction}
Convolutional neural networks have become a foundational approach for visual recognition because they learn hierarchical representations directly from image data \cite{lecun2015deep}. Their predictive strength, however, does not by itself determine whether a model is suitable for a constrained deployment context. Parameter storage and computational demand also shape whether an architecture is practical. Consequently, lightweight model selection is a multidimensional problem rather than a ranking based only on classification accuracy.

A large family of compact architectures has been developed to address these constraints. SqueezeNet reduces parameter storage through Fire modules \cite{iandola2016squeezenet}. MobileNetV2 uses inverted residual blocks and linear bottlenecks \cite{sandler2018mobilenetv2}, while MobileNetV3 combines mobile oriented search with squeeze and excitation mechanisms and specialized nonlinearities \cite{howard2019mobilenetv3}. ShuffleNetV2 explicitly considers practical runtime factors such as memory access and operator fragmentation \cite{ma2018shufflenetv2}. EfficientNet introduces compound scaling across depth, width, and input resolution \cite{tan2019efficientnet}, and EfficientNetV2 extends this direction with fused mobile blocks and training aware design \cite{tan2021efficientnetv2}. ResNet18 remains a useful conventional baseline because its residual structure and dense operators are widely supported by modern accelerators \cite{he2016resnet}.

The existence of many compact architectures creates two recurring evaluation problems. First, published comparisons often combine results produced with different data processing, optimization settings, input sizes, or pretrained checkpoints. Such comparisons can confound architectural effects with experimental choices. Second, commonly reported efficiency indicators answer different questions. Parameter count describes weight storage, while multiply accumulate operations estimate theoretical arithmetic demand. Neither measure alone determines deployment behavior, but both remain useful when they are reported consistently and interpreted alongside predictive performance.

Recent benchmark studies have examined pretrained backbones across broad task collections, domains, or resource constrained settings. These studies show that model rankings depend on the target data, training strategy, and deployment objective rather than on ImageNet accuracy alone \cite{goldblum2023battle,pegeot2023constraints,jeevan2024backbone,guerin2025vibes}. The present work takes a narrower and more controlled approach. It focuses on seven established CNNs available through one model library, applies one downstream protocol to three progressively larger classification label spaces, and evaluates two representative architectures under both ImageNet pretrained and random initialization.

Initialization is therefore treated as a separate experimental question. ImageNet pretrained weights often improve downstream accuracy and reduce the amount of task specific optimization required \cite{kornblith2019transfer}. Random initialization can become competitive when sufficiently large datasets and long schedules are available \cite{he2019rethinking}. The experiment in this paper is consequently framed as a matched fixed protocol comparison rather than a claim about the maximum attainable performance from random initialization.

The study is organized around three research questions:

\begin{enumerate}[label=\textbf{RQ\arabic*:}]
    \item How do established lightweight CNNs compare in top-1 accuracy, macro F1, and top-5 accuracy under one common fine tuning protocol?
    \item Which models provide favorable tradeoffs among predictive performance, parameter storage, and theoretical computation?
    \item How strongly does ImageNet pretrained initialization affect EfficientNet-B0 and MobileNetV3-Small under matched training settings?
\end{enumerate}

The main contributions are as follows:

\begin{itemize}
    \item A standardized benchmark of seven established lightweight CNNs across CIFAR-10, CIFAR-100, and Tiny ImageNet.
    \item A joint analysis of predictive quality, parameter storage, and theoretical computational demand, including an empirical accuracy and size Pareto analysis.
    \item A matched pretrained and random initialization study using MobileNetV3-Small and EfficientNet-B0, which represent distinct compact model regimes.
\end{itemize}

The study contributes a focused comparison that combines a shared training recipe, three common classification benchmarks, multiple resource indicators, and an initialization analysis involving two architectures.

\section{Related Work}
\subsection{Lightweight CNN design}
Compact CNNs use several distinct strategies to reduce storage or arithmetic demand. SqueezeNet relies on Fire modules to minimize parameter count \cite{iandola2016squeezenet}. MobileNetV2 uses depthwise convolution, inverted residual blocks, and linear bottlenecks \cite{sandler2018mobilenetv2}. MobileNetV3 extends this family through hardware aware search, squeeze and excitation, and specialized nonlinearities \cite{howard2019mobilenetv3}. ShuffleNetV2 emphasizes channel split and shuffle operations together with practical design rules for efficient execution \cite{ma2018shufflenetv2}. EfficientNet scales depth, width, and resolution jointly \cite{tan2019efficientnet}, while EfficientNetV2 introduces fused mobile blocks and training aware search \cite{tan2021efficientnetv2}. ResNet18 provides a conventional residual baseline built from dense convolutions \cite{he2016resnet}.

These architectures were proposed under different objectives and training conditions. Their original ImageNet results therefore do not by themselves establish how they compare after adaptation to smaller target datasets. A controlled downstream benchmark is useful because it reduces differences caused by preprocessing, optimization, and checkpoint selection.

\subsection{Backbone benchmarking and resource aware evaluation}
Large scale backbone benchmarks have shown that model selection is task dependent. Goldblum et al. evaluated more than twenty pretrained backbones across classification, detection, segmentation, retrieval, and distribution shift settings, including randomly initialized baselines \cite{goldblum2023battle}. Their study provides broad conclusions across architectures, pretraining methods, and source datasets. Jeevan and Sethi instead focused on resource efficient pretrained backbones across natural, medical, astronomical, plant, and remote sensing datasets under consistent fine tuning settings \cite{jeevan2024backbone}. They found that rankings change across domains and data regimes.

More recent work has moved from generic ranking toward target specific model selection. Guerin et al. formalized efficient backbone selection for low data image classification and showed that a short dataset specific search can outperform generic recommendations \cite{guerin2025vibes}. Resource constrained evaluation has also been studied through direct budgeted challenges. Tiwari et al. summarized ICCV 2023 challenges that constrained training time, GPU memory, and inference time, showing that accuracy and efficiency depend on both architecture choice and engineering decisions \cite{tiwari2023rcv}.

The present study differs in scope from these broader benchmarks and challenges. It does not attempt to identify a universal backbone, compare pretraining algorithms, or measure device latency. Instead, it provides a compact reference for seven established CNNs under one full fine tuning recipe on three widely used natural image datasets. It further adds a matched initialization comparison for two architectures and reports class balanced predictive performance together with storage and GMAC measures.

\subsection{Pretraining under constrained downstream budgets}
Kornblith et al. showed that stronger ImageNet models often transfer better to downstream classification tasks \cite{kornblith2019transfer}. He et al. demonstrated that random initialization can become competitive in detection when the target data and optimization schedule are sufficiently large \cite{he2019rethinking}. Pegeot et al. studied compact architectures under constrained transfer learning and found that the effect of architecture and downstream strategy changes across low resource settings \cite{pegeot2023constraints}. These findings suggest that pretrained and random initialization comparisons must be interpreted with respect to the downstream data and optimization budget.

The current initialization experiment therefore asks a limited practical question: under one shared downstream recipe, how large is the observed advantage of ImageNet pretrained weights for one compact mobile model and one stronger efficient model? It does not estimate the best result attainable from random initialization with architecture specific tuning or longer training.

\begin{table}[H]
\centering
\caption{Position of the present study relative to recent backbone and resource aware benchmarks.}
\label{tab:relatedcomparison}
\small
\renewcommand{\arraystretch}{1.08}
\begin{tabularx}{\textwidth}{>{\raggedright\arraybackslash}p{0.18\textwidth}>{\raggedright\arraybackslash}p{0.37\textwidth}>{\raggedright\arraybackslash}X}
\toprule
Study & Main scope & Difference from the present study \\
\midrule
Goldblum et al. \cite{goldblum2023battle} & More than twenty backbones across multiple vision tasks and pretraining methods, including random initialization & Covers a much broader task and pretraining space rather than established lightweight CNNs under one compact downstream protocol \\
Pegeot et al. \cite{pegeot2023constraints} & Compact architectures transferred to six downstream tasks using different upstream data compositions, linear probing, and full fine tuning & Examines transfer mechanisms under constraints rather than storage and GMAC tradeoffs among a fixed set of public lightweight models \\
Tiwari et al. \cite{tiwari2023rcv} & Budgeted training and inference challenges with direct constraints on memory, time, and latency & Evaluates complete constrained systems and engineering strategies rather than one matched architecture benchmark \\
Jeevan and Sethi \cite{jeevan2024backbone} & Lightweight pretrained backbones across diverse application domains and low data settings & Provides broader domain coverage without the same three dataset progression or matched two model initialization study \\
Guerin et al. \cite{guerin2025vibes} & Efficient target specific backbone selection for image classification in low data settings & Proposes a model selection strategy rather than a fixed controlled benchmark of a defined CNN set \\
Present study & Seven established CNNs on CIFAR-10, CIFAR-100, and Tiny ImageNet, with predictive metrics, parameter storage, GMACs, and matched initialization & Provides a focused seven model comparison under one downstream recipe across three datasets \\
\bottomrule
\end{tabularx}
\end{table}

\section{Experimental Methodology}
\subsection{Datasets and evaluation splits}
Three balanced image classification benchmarks are used. CIFAR-10 contains 50,000 training images and 10,000 test images across 10 classes. CIFAR-100 uses the same total counts with 100 classes \cite{krizhevsky2009cifar}. Tiny ImageNet contains 200 classes, with 500 training images and 50 validation images per class at a native resolution of 64 by 64 pixels \cite{le2015tiny}. Since the official Tiny ImageNet test labels are not publicly distributed, the labeled validation set is used as the test set.

For each dataset, 10\% of the official training images are selected as a validation set through a stratified split with split seed 42. This produces 45,000 training and 5,000 validation images for each CIFAR dataset, and 90,000 training and 10,000 validation images for Tiny ImageNet. The official CIFAR test sets and the Tiny ImageNet labeled validation set are not used for model selection.

\begin{table}[H]
\centering
\caption{Dataset composition used in the benchmark.}
\label{tab:datasets}
\begin{tabular}{lrrrr}
\toprule
Dataset & Classes & Train & Validation & Test \\
\midrule
CIFAR-10 & 10 & 45,000 & 5,000 & 10,000 \\
CIFAR-100 & 100 & 45,000 & 5,000 & 10,000 \\
Tiny ImageNet & 200 & 90,000 & 10,000 & 10,000 \\
\bottomrule
\end{tabular}
\end{table}

\subsection{Evaluated architectures}
Seven architectures are selected to represent different approaches to compact CNN design. All pretrained models are instantiated from torchvision 0.26.0 using publicly available ImageNet-1K weight variants. The original classifier is replaced with a task specific output layer, and the entire network is fine tuned.

\begin{table}[H]
\centering
\caption{Architectures and the design roles represented in the benchmark.}
\label{tab:models}
\begin{tabularx}{\textwidth}{l l X}
\toprule
Model & Design family & Primary role in the comparison \\
\midrule
SqueezeNet1.1 & Fire modules & Very small parameter storage \\
ShuffleNetV2 x1.0 & Channel split and shuffle & Hardware aware lightweight design \\
MobileNetV2 & Inverted residual blocks & Established mobile baseline \\
MobileNetV3-Small & Searched mobile blocks & Compact mobile accuracy and computation \\
ResNet18 & Residual blocks & Conventional dense convolution baseline \\
EfficientNet-B0 & Compound scaling & Balanced accuracy and efficiency \\
EfficientNetV2-S & Fused mobile blocks and scaling & Accuracy oriented efficient architecture \\
\bottomrule
\end{tabularx}
\end{table}

\subsection{Preprocessing and augmentation}
All images are processed at 224 by 224 pixels to enforce a common input resolution across architectures. During training, a random resized crop is applied with a scale range of 0.80 to 1.00 and an aspect ratio range of 0.90 to 1.10 using bicubic interpolation. Images are horizontally flipped with probability 0.5. Color jitter uses brightness, contrast, and saturation factors of 0.15 and a hue factor of 0.05. Each image is converted to a tensor and normalized with the ImageNet channel means $(0.485, 0.456, 0.406)$ and standard deviations $(0.229, 0.224, 0.225)$.

Validation and test images are resized directly to 224 by 224 pixels with bicubic interpolation, converted to tensors, and normalized with the same statistics. No random augmentation is used during evaluation.

\subsection{Training protocol}
The benchmark is implemented in PyTorch \cite{paszke2019pytorch}. All experiments use training seed 42 and split seed 42. Python, NumPy, PyTorch, and CUDA random number generators are seeded, and deterministic cuDNN behavior is enabled. The models are optimized with AdamW \cite{loshchilov2019adamw}, an initial learning rate of $3\times10^{-4}$, weight decay of $10^{-4}$, and label smoothing of 0.1 \cite{szegedy2016inception}. A cosine annealing schedule decreases the learning rate over a maximum of 20 epochs \cite{loshchilov2017sgdr}. Automatic mixed precision is enabled, and the gradient norm is clipped to 1.0.

The training batch size is 32 and the validation and test batch size is 64. Early stopping uses a patience of five epochs. The primary checkpoint criterion is validation accuracy, with validation macro F1 used to resolve equal accuracy. The best checkpoint is evaluated once on the test set. Most runs complete the full 20 epochs. ShuffleNetV2 x1.0 on CIFAR-10 stops after 12 epochs because its validation score does not improve for five consecutive epochs.

\begin{table}[H]
\centering
\caption{Common training configuration.}
\label{tab:protocol}
\begin{tabular}{ll}
\toprule
Setting & Value \\
\midrule
Training seed and split seed & 42 \\
Input resolution & 224 by 224 \\
Maximum epochs & 20 \\
Training batch size & 32 \\
Evaluation batch size & 64 \\
Optimizer & AdamW \\
Initial learning rate & $3\times10^{-4}$ \\
Weight decay & $10^{-4}$ \\
Label smoothing & 0.1 \\
Scheduler & Cosine annealing \\
Early stopping patience & 5 epochs \\
Gradient clipping & 1.0 \\
Precision & Automatic mixed precision \\
Checkpoint selection & Validation accuracy, then macro F1 \\
\bottomrule
\end{tabular}
\end{table}

\subsection{Pretrained and random initialization study}
EfficientNet-B0 and MobileNetV3-Small are additionally trained from random initialization. These models represent distinct computational regimes: EfficientNet-B0 has approximately 4.0 to 4.3 million task specific parameters, while MobileNetV3-Small has approximately 1.5 to 1.7 million. The scratch trained experiments use the same data split, augmentation, optimizer, learning rate, scheduler, batch sizes, label smoothing, mixed precision setting, and maximum epoch budget as the corresponding pretrained runs. The comparison therefore holds the downstream protocol constant while varying initialization.

\subsection{Evaluation metrics}
Predictive performance is measured with top-1 accuracy, macro F1, and top-5 accuracy. Macro F1 gives equal weight to each class and provides a classwise complement to overall accuracy. Top-5 accuracy is particularly informative for CIFAR-100 and Tiny ImageNet, where the number of classes is larger.

Model efficiency is described by total parameter count, FP32 parameter storage, and GMACs. FP32 storage is calculated as four bytes per parameter and is reported in mebibytes. GMACs are profiled for one input tensor of shape $1\times3\times224\times224$. Multiply accumulate counts are reported rather than doubling them into a nominal floating point operation estimate.

\section{Results}
\subsection{Predictive performance of pretrained models}
\Cref{tab:mainresults} presents top-1 accuracy and macro F1 for the seven pretrained models. EfficientNetV2-S records the highest observed top-1 accuracy on every dataset. It obtains 97.57\% accuracy on CIFAR-10, 86.98\% on CIFAR-100, and 78.73\% on Tiny ImageNet. EfficientNet-B0 ranks second in all three cases and remains close to the larger model, with accuracy gaps of only 0.22, 0.85, and 0.65 percentage points.

ResNet18 performs strongly on CIFAR-10 and CIFAR-100, reaching 96.52\% and 82.53\% accuracy. MobileNetV3-Small also remains competitive across all three datasets while operating with the smallest GMAC count in the benchmark. This combination makes it especially relevant for ultra low resource settings. ShuffleNetV2 x1.0 provides moderate predictive performance, while SqueezeNet1.1 produces the lowest accuracy and macro F1 on all three datasets.

\begin{table}[H]
\centering
\caption{Pretrained test accuracy and macro F1. Best values are shown in bold and second best values are underlined.}
\label{tab:mainresults}
\small
\begin{tabular}{l cc cc cc}
\toprule
\multirow{2}{*}{Model} & \multicolumn{2}{c}{CIFAR-10} & \multicolumn{2}{c}{CIFAR-100} & \multicolumn{2}{c}{Tiny ImageNet} \\
\cmidrule(lr){2-3}\cmidrule(lr){4-5}\cmidrule(lr){6-7}
& Acc. (\%) & Macro F1 & Acc. (\%) & Macro F1 & Acc. (\%) & Macro F1 \\
\midrule
EfficientNetV2-S & \textbf{97.57} & \textbf{0.9757} & \textbf{86.98} & \textbf{0.8697} & \textbf{78.73} & \textbf{0.7866} \\
EfficientNet-B0 & \underline{97.35} & \underline{0.9735} & \underline{86.13} & \underline{0.8615} & \underline{78.08} & \underline{0.7801} \\
ResNet18 & 96.52 & 0.9652 & 82.53 & 0.8250 & 68.22 & 0.6815 \\
MobileNetV3-Small & 96.12 & 0.9611 & 82.43 & 0.8239 & 70.95 & 0.7093 \\
MobileNetV2 & 95.99 & 0.9599 & 81.45 & 0.8144 & 71.41 & 0.7126 \\
ShuffleNetV2 x1.0 & 94.78 & 0.9477 & 80.38 & 0.8036 & 69.35 & 0.6925 \\
SqueezeNet1.1 & 93.73 & 0.9371 & 73.34 & 0.7330 & 60.41 & 0.6030 \\
\bottomrule
\end{tabular}
\end{table}

\begin{figure}[H]
\centering
\includegraphics[width=\textwidth]{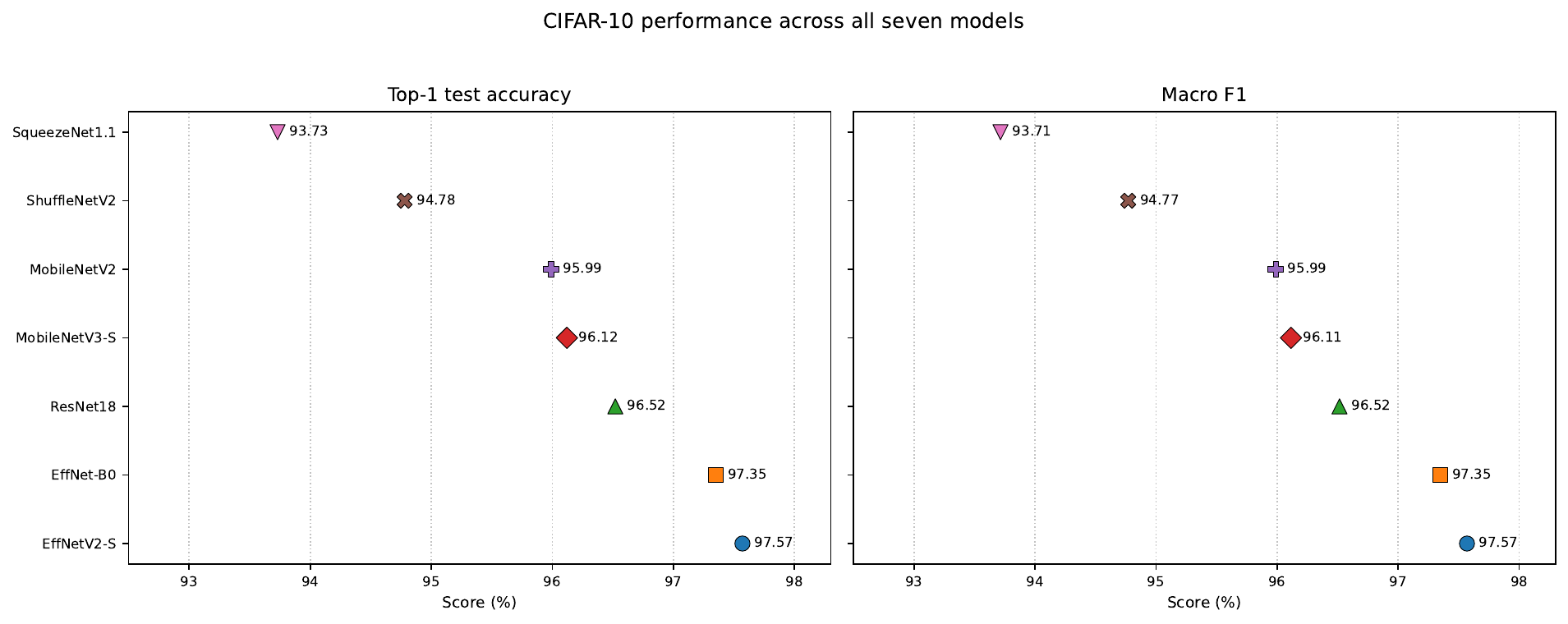}
\caption{CIFAR-10 top-1 accuracy and macro F1 across the seven models.}
\label{fig:cifar10performance}
\end{figure}

\begin{figure}[H]
\centering
\includegraphics[width=\textwidth]{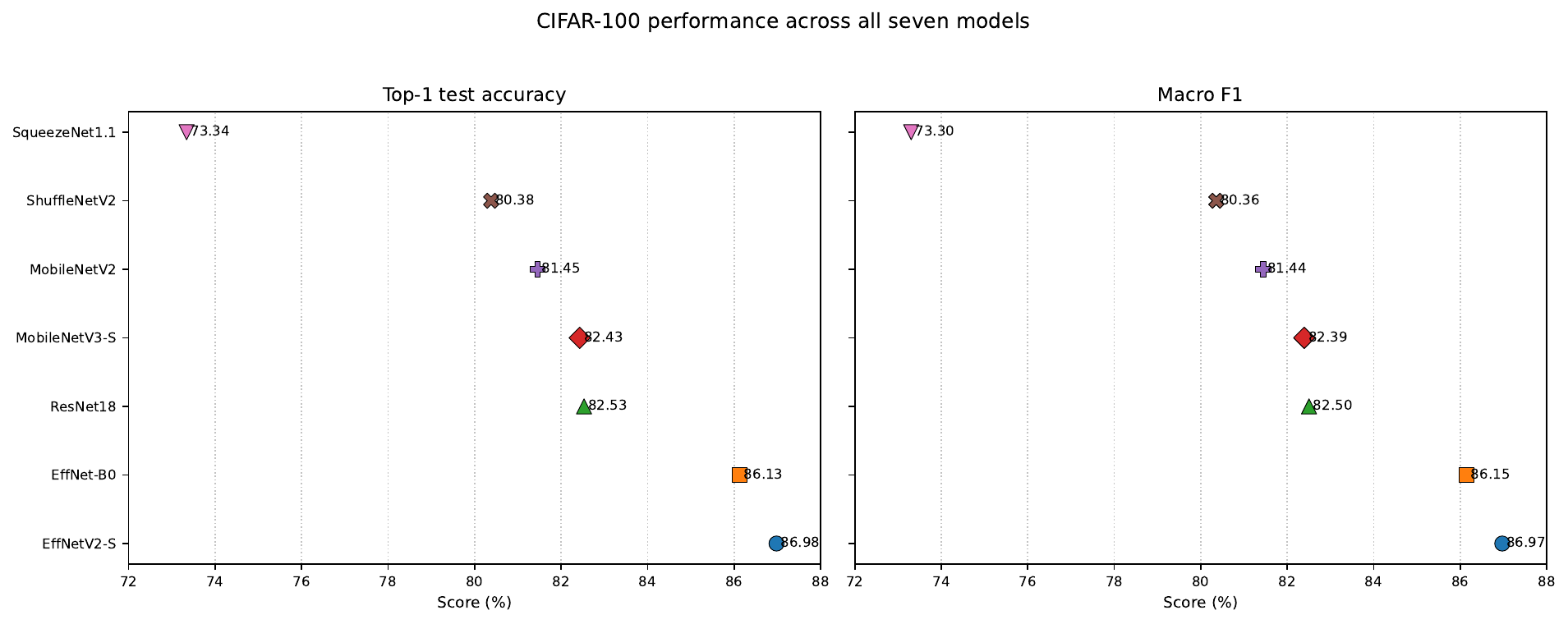}
\caption{CIFAR-100 top-1 accuracy and macro F1 across the seven models.}
\label{fig:cifar100performance}
\end{figure}

\begin{figure}[H]
\centering
\includegraphics[width=\textwidth]{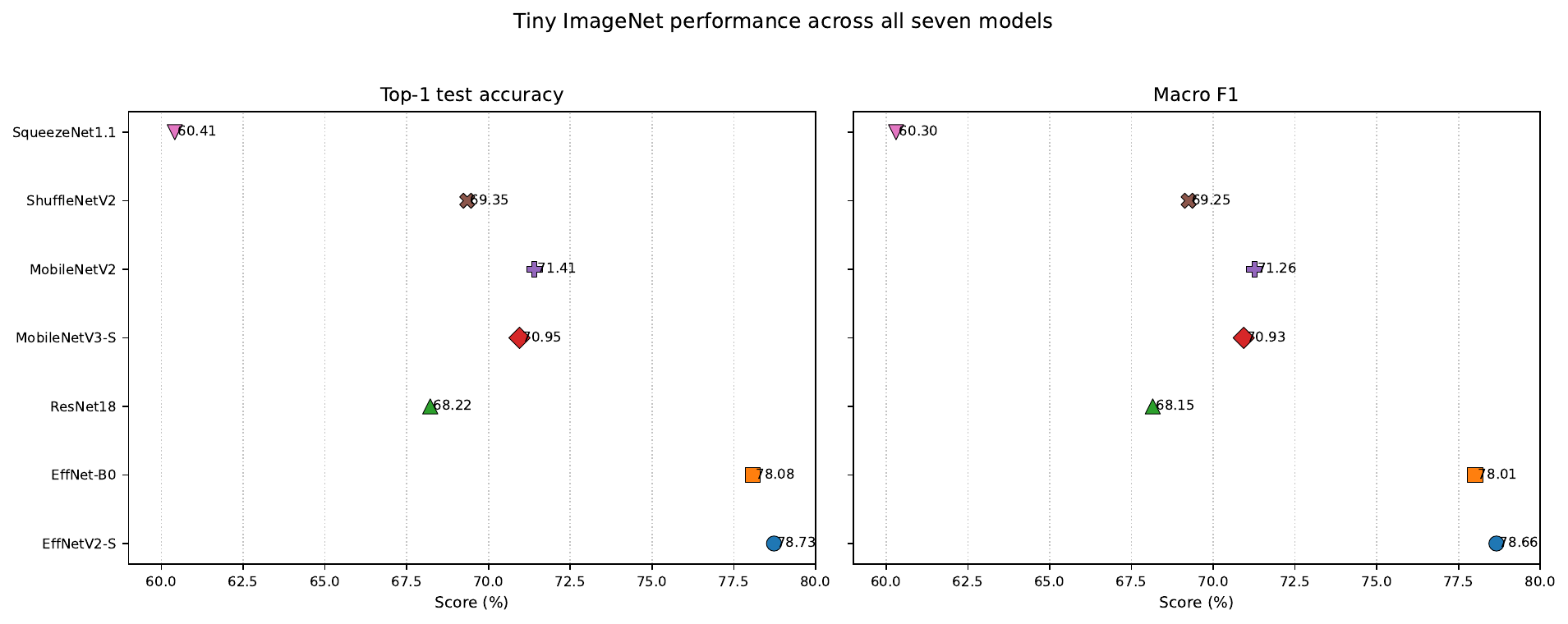}
\caption{Tiny ImageNet top-1 accuracy and macro F1 across the seven models.}
\label{fig:tinyperformance}
\end{figure}

Top-5 performance is reported in \cref{tab:top5}. EfficientNet-B0 records the highest top-5 accuracy on CIFAR-100 at 97.29\%. EfficientNetV2-S remains close at 96.69\%. On Tiny ImageNet, EfficientNet-B0 again ranks first at 92.42\%, followed by EfficientNetV2-S at 91.79\%. EfficientNet-B0 achieves the highest observed top-5 accuracy on both higher class count datasets, while EfficientNetV2-S achieves the highest observed top-1 accuracy.

\begin{table}[H]
\centering
\caption{Pretrained top-5 accuracy on the higher class count datasets.}
\label{tab:top5}
\begin{tabular}{lcc}
\toprule
Model & CIFAR-100 (\%) & Tiny ImageNet (\%) \\
\midrule
EfficientNetV2-S & 96.69 & 91.79 \\
EfficientNet-B0 & \textbf{97.29} & \textbf{92.42} \\
ResNet18 & 95.42 & 86.33 \\
MobileNetV3-Small & 96.48 & 89.32 \\
MobileNetV2 & 95.92 & 89.57 \\
ShuffleNetV2 x1.0 & 96.19 & 88.69 \\
SqueezeNet1.1 & 93.65 & 83.13 \\
\bottomrule
\end{tabular}
\end{table}

\subsection{Parameter storage and computational demand}
\Cref{tab:efficiency} reports efficiency measures for the Tiny ImageNet variants, which have the largest classifiers among the three tasks. SqueezeNet1.1 is the smallest model at 0.825 million parameters and 3.15 MiB of FP32 parameter storage. ShuffleNetV2 x1.0 and MobileNetV3-Small are also compact at 1.46 million and 1.72 million parameters. MobileNetV3-Small has the lowest theoretical computation at 0.0617 GMACs.

EfficientNetV2-S is the largest and most computationally demanding model in the benchmark. It uses 20.43 million parameters and 2.90 GMACs. EfficientNet-B0 remains close to its predictive performance while reducing parameter count by a factor of 4.79 and GMACs by a factor of 7.00. This places EfficientNet-B0 on the empirical accuracy and size Pareto frontier and gives it a favorable general balance in the evaluated set.

MobileNetV3-Small occupies the ultra low resource end of the comparison. Relative to EfficientNet-B0 on the Tiny ImageNet configuration, it uses about 40\% as many parameters and 15\% as many GMACs. Its predictive accuracy remains competitive despite the smaller footprint, making it a strong candidate when storage and theoretical computation are dominant constraints.

\begin{table}[H]
\centering
\caption{Efficiency measures for the Tiny ImageNet model variants.}
\label{tab:efficiency}
\small
\begin{tabular}{lrrr}
\toprule
Model & Params (M) & FP32 (MiB) & GMACs \\
\midrule
EfficientNetV2-S & 20.434 & 77.948 & 2.9009 \\
EfficientNet-B0 & 4.264 & 16.265 & 0.4141 \\
ResNet18 & 11.279 & 43.026 & 1.8236 \\
MobileNetV3-Small & 1.723 & 6.572 & \textbf{0.0617} \\
MobileNetV2 & 2.480 & 9.461 & 0.3265 \\
ShuffleNetV2 x1.0 & 1.459 & 5.564 & 0.1519 \\
SqueezeNet1.1 & \textbf{0.825} & \textbf{3.147} & 0.2800 \\
\bottomrule
\end{tabular}
\end{table}

\begin{figure}[H]
\centering
\includegraphics[width=\textwidth]{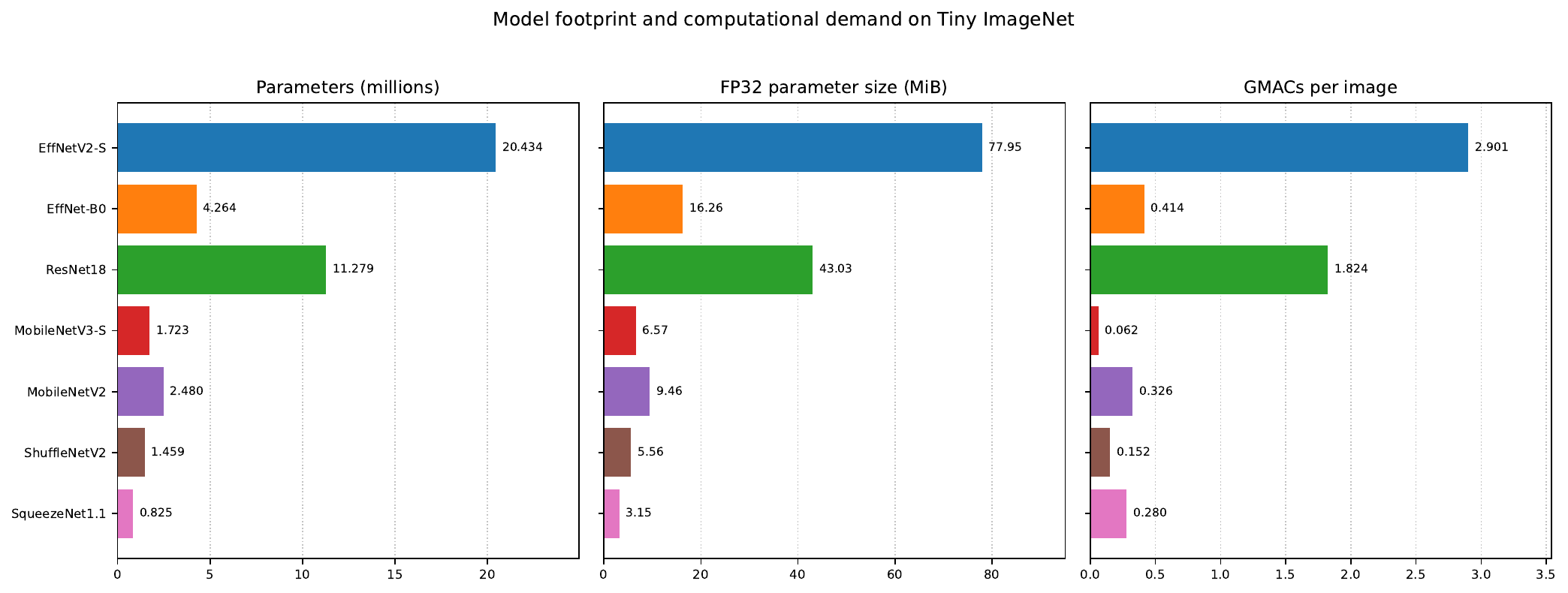}
\caption{Parameter count, FP32 parameter size, and GMACs for the Tiny ImageNet variants. Exact values are displayed beside the bars.}
\label{fig:efficiencyprofile}
\end{figure}

\begin{figure}[H]
\centering
\includegraphics[width=\textwidth]{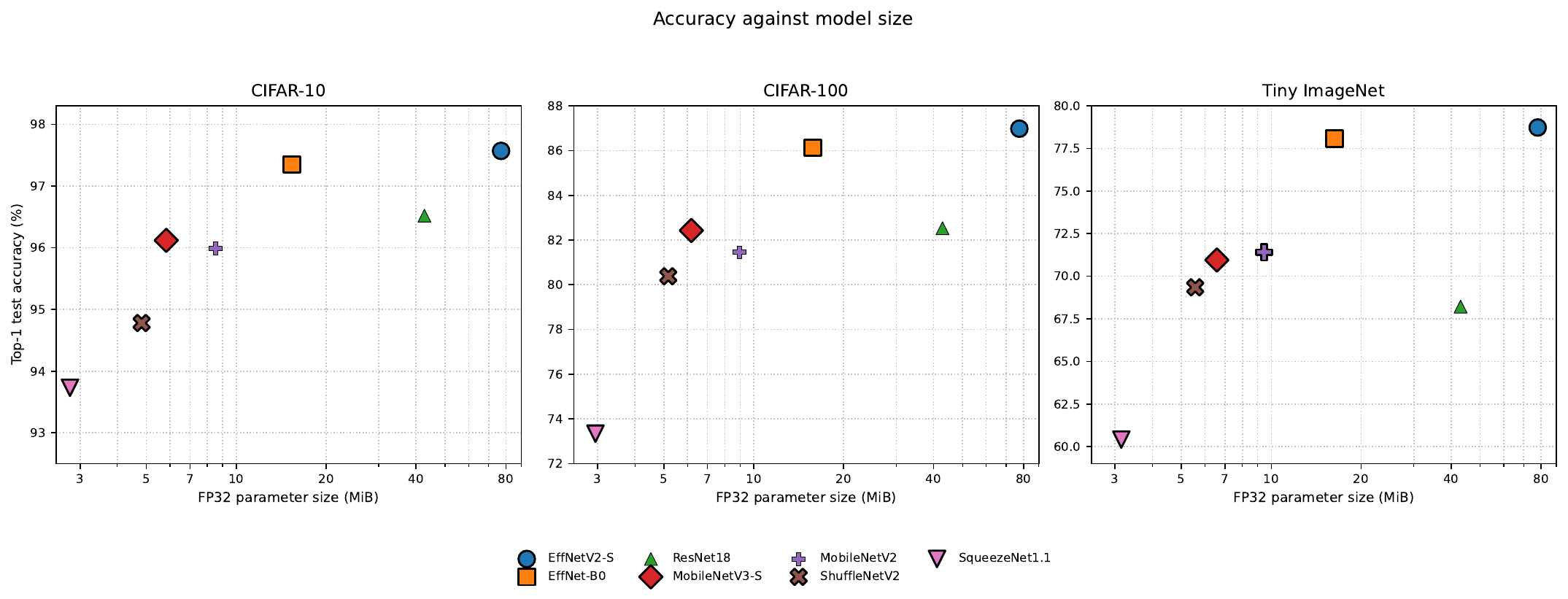}
\caption{Top-1 accuracy against FP32 parameter size. Larger outlined markers indicate Pareto efficient models, meaning that no evaluated alternative is both smaller and more accurate.}
\label{fig:accuracyvssize}
\end{figure}

\subsection{Impact of pretrained initialization}
\Cref{tab:pretraining} compares pretrained and randomly initialized EfficientNet-B0 and MobileNetV3-Small models under matched training settings. Pretrained initialization improves every predictive metric for both architectures and all datasets.

The randomly initialized models remain comparatively strong on CIFAR-10, but the observed gaps are much larger on CIFAR-100 and Tiny ImageNet. Because the datasets differ in class count, image source, and relationship to ImageNet, the experiment does not isolate a single cause for this pattern. The exact accuracy and macro F1 differences are reported in \cref{tab:pretraining}.

\begin{table}[H]
\centering
\caption{Matched protocol comparison of pretrained and random initialization. Acc. gap is measured in percentage points and F1 gap is an absolute difference.}
\label{tab:pretraining}
\small
\resizebox{\textwidth}{!}{%
\begin{tabular}{llrrrrrr}
\toprule
Model & Dataset & Pre. Acc. & Random Acc. & Acc. gap & Pre. F1 & Random F1 & F1 gap \\
\midrule
\multirow{3}{*}{EfficientNet-B0}
& CIFAR-10 & 97.35 & 91.97 & 5.38 & 0.9735 & 0.9193 & 0.0542 \\
& CIFAR-100 & 86.13 & 71.23 & 14.90 & 0.8615 & 0.7117 & 0.1497 \\
& Tiny ImageNet & 78.08 & 57.43 & 20.65 & 0.7801 & 0.5706 & 0.2095 \\
\midrule
\multirow{3}{*}{MobileNetV3-Small}
& CIFAR-10 & 96.12 & 88.09 & 8.03 & 0.9611 & 0.8801 & 0.0811 \\
& CIFAR-100 & 82.43 & 60.30 & 22.13 & 0.8239 & 0.6002 & 0.2237 \\
& Tiny ImageNet & 70.95 & 47.75 & 23.20 & 0.7093 & 0.4721 & 0.2372 \\
\bottomrule
\end{tabular}%
}
\end{table}

\begin{figure}[H]
\centering
\includegraphics[width=0.82\textwidth]{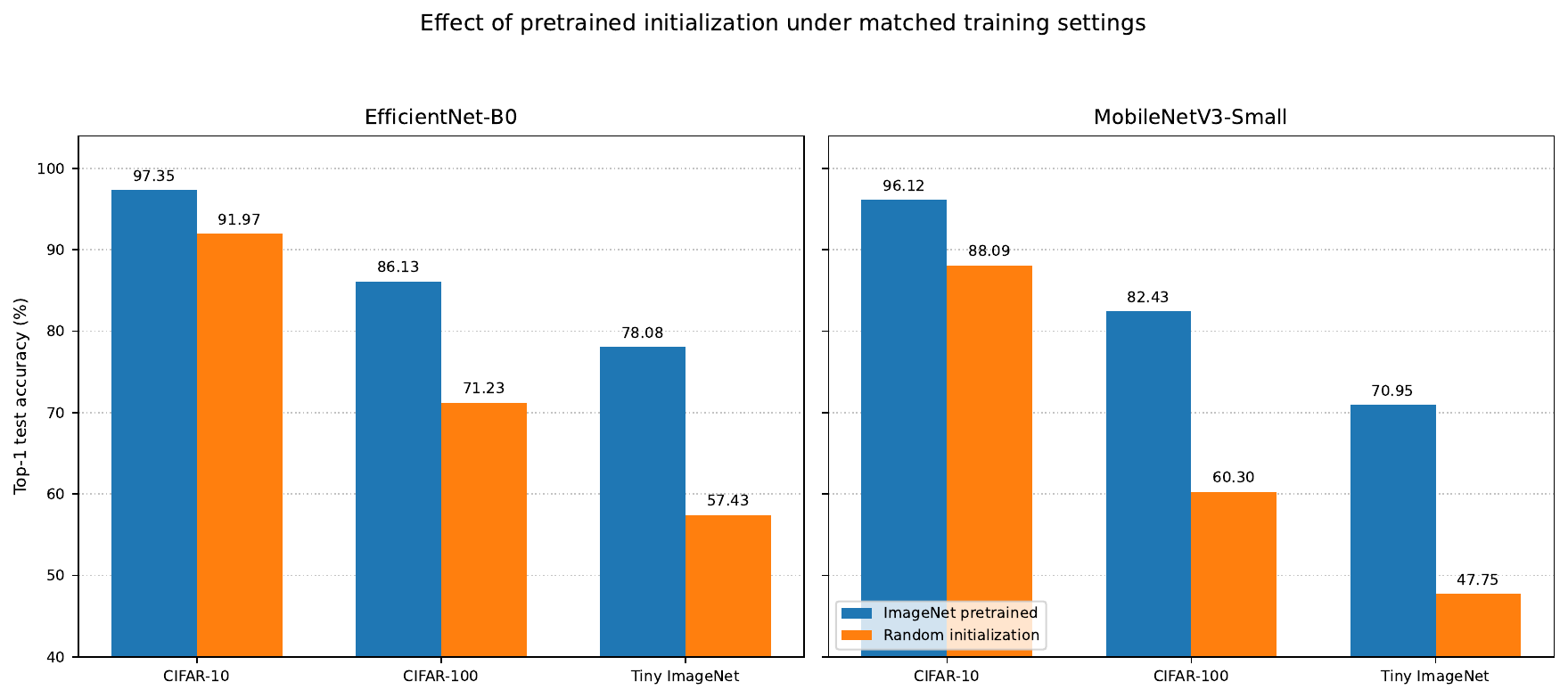}
\caption{Top-1 accuracy for ImageNet pretrained and randomly initialized models under matched training settings.}
\label{fig:pretraining}
\end{figure}

Under the fixed protocol, both randomly initialized models exceed 88\% accuracy on CIFAR-10, while much larger gaps are observed on CIFAR-100 and Tiny ImageNet. The result is consistent with pretrained representations providing useful features at the start of downstream optimization, but it should not be interpreted as isolating dataset complexity as the sole cause of the differences.

Top-5 results show a similar pattern. For EfficientNet-B0, the pretrained advantage is 0.42, 5.43, and 11.90 percentage points across CIFAR-10, CIFAR-100, and Tiny ImageNet. For MobileNetV3-Small, the corresponding differences are 0.47, 9.74, and 16.10 points.

\section{Discussion}
\subsection{Model choices under different resource constraints}
The results support three practical model profiles. EfficientNetV2-S is the accuracy oriented option because it records the highest observed top-1 accuracy on all three datasets, although it also has the largest storage and GMAC requirements. EfficientNet-B0 offers a favorable general balance. It remains close to EfficientNetV2-S while requiring only about 21\% of its parameters and 14\% of its GMACs on the Tiny ImageNet configuration.

MobileNetV3-Small is a strong ultra low resource candidate. It uses about 40\% as many parameters and 15\% as many GMACs as EfficientNet-B0, while retaining competitive predictive performance. Its position on the empirical accuracy and size Pareto frontier for all three datasets supports this interpretation. SqueezeNet1.1 provides the minimum storage, but its larger accuracy reduction shows that minimum parameter count alone does not guarantee a favorable tradeoff. ShuffleNetV2 x1.0 also occupies the Pareto frontier, although MobileNetV3-Small provides higher observed accuracy at a modest increase in storage.

The Pareto analysis is descriptive rather than a single numerical ranking. A model is treated as Pareto efficient when no evaluated alternative is both smaller and more accurate. The result therefore identifies useful choices under different storage budgets without assigning subjective weights to accuracy, size, and GMACs.

\subsection{Effect of pretrained initialization}
ImageNet pretrained weights produce substantially higher observed accuracy and macro F1 under the matched downstream protocol. The differences are modest on CIFAR-10 relative to the gaps on CIFAR-100 and Tiny ImageNet. Since the datasets differ in several respects, including class count, native resolution, visual content, and relationship to ImageNet, the experiment does not isolate one causal factor.

The comparison also does not show that random initialization cannot approach the pretrained models with longer schedules, different learning rates, warmup, or architecture specific regularization. Prior work shows that random initialization can become competitive in suitable regimes \cite{he2019rethinking}. The supported conclusion is narrower: ImageNet pretrained initialization provides a substantial optimization and predictive advantage for these two architectures under the fixed protocol.

\subsection{Relation to broader backbone benchmarks}
The findings agree with recent evidence that model choice depends on the target setting and that no backbone dominates every objective \cite{goldblum2023battle,jeevan2024backbone,guerin2025vibes}. The present results add a focused comparison among established compact CNNs. EfficientNetV2-S leads in observed top-1 accuracy, EfficientNet-B0 provides a favorable middle ground, and MobileNetV3-Small offers a strong low resource profile. These conclusions are specific to the evaluated checkpoints, preprocessing, datasets, and resource indicators.

\section{Limitations}
This study has several limitations. First, each configuration is evaluated with one training seed. The numerical rankings are point estimates without confidence intervals, and small differences should not be interpreted as definitive evidence of stable superiority. In particular, observed gaps below one percentage point may change across repeated runs.

Second, the fixed downstream recipe supports a matched comparison but is not equally optimized for pretrained and randomly initialized models. Random initialization may benefit from longer schedules, larger learning rates, warmup, or different regularization. The initialization results therefore describe the fixed protocol rather than the best attainable random initialization performance.

Third, the public torchvision checkpoints were produced with architecture specific ImageNet training recipes. Downstream differences cannot be attributed only to network topology. The study compares publicly available pretrained model instances, which is useful for practical selection but not a controlled causal test of architecture alone.

Fourth, Tiny ImageNet is derived from ImageNet and shares its class taxonomy \cite{deng2009imagenet,le2015tiny}. Its experiment therefore represents adaptation from ImageNet pretraining to a reduced resolution subset with related labels rather than transfer to a fully independent visual domain. The pretraining advantage on Tiny ImageNet should be interpreted with this relationship in mind.

Fifth, all images are resized to 224 by 224 pixels. This preserves compatibility with ImageNet pretrained models but increases the spatial size of CIFAR and Tiny ImageNet images without adding native detail. Alternative input resolutions may change both predictive and computational tradeoffs. Finally, the paper reports parameter storage and GMACs rather than complete deployment efficiency. Device latency, energy use, runtime memory, quantization, pruning, robustness, and calibration are not measured.

\section{Conclusion}
This paper presents a standardized comparison of seven lightweight CNNs across CIFAR-10, CIFAR-100, and Tiny ImageNet. EfficientNetV2-S records the highest observed top-1 accuracy on every dataset. EfficientNet-B0 follows closely while reducing parameter count and GMACs substantially, giving it a favorable general balance in the evaluated set.

MobileNetV3-Small provides a strong ultra low resource option. It uses about 40\% as many parameters and 15\% as many GMACs as EfficientNet-B0 on the largest classifier configuration while retaining competitive predictive performance. It also lies on the empirical accuracy and size Pareto frontier across all three datasets. SqueezeNet1.1 requires the least storage, but its larger predictive loss limits its usefulness as task difficulty increases.

The matched initialization study shows that ImageNet pretrained weights provide a substantial advantage under the fixed downstream recipe. The observed gaps are much larger on CIFAR-100 and Tiny ImageNet than on CIFAR-10, although the datasets differ in several ways and the experiment does not isolate a single cause. Overall, the benchmark supports EfficientNetV2-S as the accuracy oriented option, EfficientNet-B0 as a favorable balanced option, and MobileNetV3-Small as a strong candidate for severe storage and computation constraints.

\end{document}